\def\year{2022}\relax
\documentclass[letterpaper]{article} 
\usepackage{aaai22}  
\usepackage{times}  
\usepackage{helvet}  
\usepackage{courier}  
\usepackage[hyphens]{url}  
\usepackage{graphicx} 
\usepackage{natbib}  
\usepackage{caption} 
\DeclareCaptionStyle{ruled}{labelfont=normalfont,labelsep=colon,strut=off} 
\usepackage{algorithm}
\usepackage{algorithmic}

\usepackage{newfloat}
\usepackage{listings}
\floatstyle{ruled}
\newfloat{listing}{tb}{lst}{}
\floatname{listing}{Listing}
\usepackage[labelfont=bf,list=true,font=footnotesize, margin=2pt]{subcaption}
\usepackage[labelfont=bf, font=footnotesize, margin=0pt]{caption}

\usepackage{epsfig}
\usepackage{amsmath}
\usepackage{xcolor}
\usepackage{amssymb, mathtools,amsfonts}
\usepackage{subcaption}        
\usepackage{multirow}
\usepackage{booktabs}

\newcommand{\squishlist}{
   \begin{list}{$\bullet$}
    { \setlength{\itemsep}{0pt}      \setlength{\parsep}{3pt}
      \setlength{\topsep}{3pt}       \setlength{\partopsep}{0pt}
      \setlength{\leftmargin}{1.0em} \setlength{\labelwidth}{1em}
      \setlength{\labelsep}{0.5em} } }
\newcommand{\squishend}{
    \end{list} }
\newcommand{\PP}[1]{\paragraph*{\bf #1}}

\title{Context-Aware Transfer Attacks for Object Detection}
\author {
    Zikui Cai\textsuperscript{\rm 1}, 
    Xinxin Xie\textsuperscript{\rm 2}, 
    Shasha Li\textsuperscript{\rm 2}, 
    Mingjun Yin\textsuperscript{\rm 2}, 
    Chengyu Song\textsuperscript{\rm 2},  \\
    Srikanth V. Krishnamurthy\textsuperscript{\rm 2}, 
    Amit K. Roy-Chowdhury\textsuperscript{\rm 1}, 
    M. Salman Asif\textsuperscript{\rm 1}
}
\affiliations {
    \textsuperscript{\rm 1} Electrical and Computer Engineering, University of California Riverside\\
    \textsuperscript{\rm 2} Computer Science and  Engineering, University of California Riverside\\
    \texttt{\{zcai032,xxie039,sli057,myin013\}@ucr.edu,\\ \{csong,krish\}@cs.ucr.edu, \{amitrc,sasif\}@ece.ucr.edu}
}

\begin{document}

\maketitle

\begin{abstract}
Blackbox transfer attacks for image classifiers have been extensively studied in recent years. In contrast, little progress has been made on transfer attacks for object detectors. Object detectors take a holistic view of the image and the detection of one object (or lack thereof) often depends on other objects in the scene. This makes such detectors inherently context-aware and adversarial attacks in this space are more challenging than those targeting image classifiers. In this paper, we present a new approach to generate context-aware attacks for object detectors. We show that by using co-occurrence of objects and their relative locations and sizes as context information, we can successfully generate targeted mis-categorization attacks that achieve higher transfer success rates on blackbox object detectors than the state-of-the-art. We test our approach on a variety of object detectors with images from PASCAL VOC and MS COCO datasets and demonstrate up to $20$ percentage points improvement in performance compared to the other state-of-the-art methods. 
\end{abstract}

\section{Introduction}

Generating adversarial attacks (and defending against such attacks) has recently gained a lot of attention. An overwhelming majority of work in these areas have considered cases when images contain one predominant object  (e.g., ImageNet~\cite{deng2009imagenet} data), and the goal is to perturb an image to change its label. In real-life situations, we usually encounter images with many objects. Object detectors take a holistic view of the image and the detection of one object (or lack thereof) depends on other objects in the scene. This is why object detectors are inherently  context-aware and adversarial attacks are more challenging than those targeting image  classifiers~\cite{goodfellow2014explaining,moosavi2016deepfool,carlini2017towards,Liu2017Delving}. 

In this paper, we focus on the problem of generating context-aware adversarial attacks on images to affect the performance of object detectors. Our approach is to craft an \textbf{attack plan} for each object, which not only perturbs a specific \textbf{victim object} to the target class, but also perturbs other objects in the image to specific labels or inserts phantom objects to enhance the holistic context consistency; these associated objects are called \textbf{helper objects}. The helpers are selected based on the \textbf{context graphs}, which capture the co-occurrence relationships and relative location and size of objects in the image. The context graphs can be learned empirically from natural image datasets. The nodes of a context graph are object classes, and each edge weight captures the co-occurrence, relative distance, and size likelihood of one object conditioned on the other. The intuition is that each class is often associated with certain classes, and unlikely to be associated with certain others.

\begin{figure*}[t]
\centering 
\includegraphics[width=0.9\linewidth]{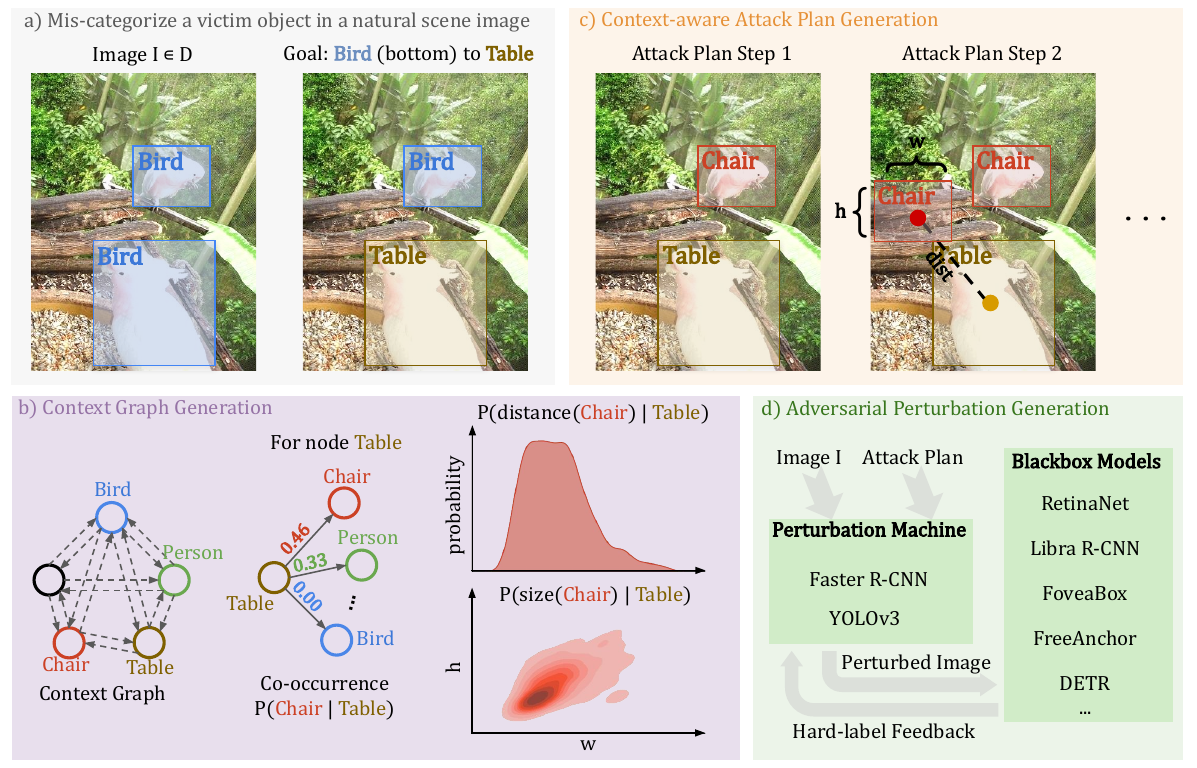}
\caption{Overview of our framework for generating the context-aware sequential attack. 
a) Given a natural image, our goal is to trick an object detector to assign the victim object a given target label (e.g., \textcolor[HTML]{3c78d8}{bird} to \textcolor[HTML]{7f6000}{table}). 
%
b) We construct a context graph that encodes the co-occurrence probability, distance, and relative size distribution relating pairs of objects (e.g., the edge from \textcolor[HTML]{7f6000}{table} to \textcolor[HTML]{cc4125}{chair} represents they co-occur with probability 0.46). 
%
c) Given the attack goal and context graph, we generate a context-aware attack plan that has a small number of steps. In each step, we assign target labels for existing objects and introduce new helper objects if needed. For example, co-occurrence of \textcolor[HTML]{cc4125}{chair} with \textcolor[HTML]{7f6000}{table} is most probable, we change the \textcolor[HTML]{3c78d8}{bird} to a \textcolor[HTML]{cc4125}{chair} for stronger context consistency (depicted in Attack Plan Step 1). We may need to add a phantom \textcolor[HTML]{cc4125}{chair} around the \textcolor[HTML]{7f6000}{table} (as depicted in Attack Plan Step 2).
%
d) Given the attack plan and the victim image, we generate perturbations using I-FGSM on the surrogate whitebox models in our perturbation machine. We test the perturbed image with the given blackbox model and based on the hard-label feedback, we either stop (when the attack is successful or when we exhaust our budget of the helper objects) or craft new attack based on the next steps and repeat the process. 
%
}
\label{fig:intro}
\end{figure*}

Our interest lies in blackbox attacks where the perturbations generated for an image are effective on a variety of detectors that may not be known during the perturbation generation process. 
The conceptual idea of our proposed approach is to generate perturbations with an ensemble of detectors and subsequently test them on an unknown detector. Such attacks are referred to as transfer attacks, and we refer to the unknown detector we seek to fool as the \textbf{victim blackbox model}. 
To achieve this goal, we propose a novel sequential strategy to generate these attacks. We sequentially add perturbations to cause the modification of the labels of the victim and helper objects, based on the co-occurrence object relation graph of the victim object. This strategy is the first to use explicit context information of an image to generate a blackbox attack plan. 
Note that the sequential strategy makes a small number of queries (2--6 in our experiments) to the blackbox detector as new helper objects are added in the attack plan. The blackbox detector provides hard labels and locations of detected objects. We use this information only as a stopping criterion for the attack generation, unlike query-based approaches \cite{Wang2020blackobject} that often need to use thousands of queries to estimate local gradients. The framework is illustrated in Figure~\ref{fig:intro}.
%

The main contributions of this paper are as follows.
\squishlist
\item This is the first work that considers co-occurrence between different object classes in complex images with multiple objects to generate transferable adversarial attacks on blackbox object detectors.  
\item We show how to generate context-aware attack plans for targeted mis-categorization attacks. The attacks generated using context-aware attack plans provide significantly better transfer success rates on blackbox detectors than those generated by methods that are agnostic to context information (an average improvement of more than 10 percentage points in fooling rate, see Table~\ref{tab:combined}). 
\item Our comprehensive evaluations also include context-aware adversarial attacks on multiple datasets using multiple object detectors. We also provide analysis on the effect of helper objects in generating successful attacks and the generalizability of contexts.
\squishend

\section{Related Work}
\PP{Context in object detection.}
The importance of context has been studied extensively to enhance visual recognition technologies \cite{strat1991context,torralba2004contextual,divvala2009empirical,galleguillos2010context,marques2011context,Yao2012WholeScene,Mottaghi2014ConWild}. Modern object detectors \cite{ren2015faster,redmon2018yolov3,carion2020end} consider holistic information in the image to locate and detect different objects, and several works explicitly utilize context information to improve the performance of object detectors \cite{bell2016inside,zhang2017relationship,chen2018context,liu2018structure,barnea2019exploring,wang2020robust}. 
Some recent papers have considered context consistency to detect adversarial attacks~\cite{li2020connecting,yin2021exploiting}, but the attack generation uses existing whitebox attack schemes that do not consider context information explicitly. To the best of our knowledge, we are the first to use context information of objects explicitly for generating attacks on object detectors for images with multiple objects.

\PP{Blackbox adversarial attacks.}
Blackbox attacks is a practical setting where the attacker can only query the model and get the output instead of having access to the model's internal parameters. Two common strategies targeting this challenging problem are transfer-based attacks and query-based attacks. Query-based attacks have high success rates but require an overwhelmingly large number (often hundreds or thousands) of queries \cite{brendel2017decision,Chen2017zoo,guo2019simple,huang2019black,Cheng2019Improving,chen2020hopskipjumpattack,li2020qeba,Wang2020blackobject}. In this paper, we explore a more stringent case where only a very small number of model calls is allowed. Several papers \cite{Papernot2017Practical,Liu2017Delving,dong2018boosting,li2020towards} have examined the phenomenon of transfer attacks where the adversarial examples generated using a surrogate network can fool an unknown network. The previous works studying transfer attacks focus on image classifiers. In this paper, we focus on object detectors, which is considered to be a much harder problem \cite{xie2017adversarial,wu2020making}. 


\PP{Attacking object detectors.} 
Almost all existing attacks on object detectors focus on whitebox setting. Some patch-based attacks \cite{Liu2019Dpatch,saha2020role} are very effective but the patches are obviously visible to observers. Some attacks such as DAG~\cite{xie2017adversarial}, RAP~\cite{li2018robust}, and CAP~\cite{Zhang2020CAA} rely on region proposal network (RPN), thus only work for proposal-based (two-stage) object detectors. Some attacks are more generic such as UAE~\cite{Wei2019Transferable} and TOG~\cite{chow2020adversarial} that work for both one-stage and two-stage models. Among them TOG is the most generic approach that can attack all different kinds of models regardless of their architectures as long as backpropagation on training loss is feasible. 
Even though some of these works have reported transfer attack results on a small set of blackbox models, since they are mainly designed for whitebox attacks, they fail to provide systematic evaluation in a realistic blackbox settings.

\section{Context-Aware Sequential Attacks}\label{method}

While algorithms in prior approaches search for adversarial examples that misclassify the victim objects only, we propose to formulate the optimization problem towards perturbing both the victim object and the ``context'' associated with the victim object. The context of an object is determined by the objects that co-exist with it. We hypothesize that the context not only plays an important role in improving classification/detection performance, but can also boost the ability to realize efficient adversarial attacks against object detectors. 

Next, we show how we compose context-aware attack plans and search for adversarial examples by sequentially solving optimization problems that are defined for a context-aware attack plan. The context-aware attack plans utilize the contextual information with regard to the co-existence of instances of different categories and their relative locations and sizes. We first describe how we represent the contextual information. Then we discuss how to compose the so-defined context-aware attack plan. Finally, we describe how we generate the adversarial examples by solving relevant optimization problems sequentially. The framework is explained in detail in Figure~\ref{fig:intro}.

\subsection{Context Modeling}
We represent a natural scene image as $I$ and the distribution of all natural images as $\mathcal{D}$. Each $I\in\mathcal{D}$ could contain one or multiple object instances. We denote the possible object categories in the distribution $\mathcal{D}$ by $C = \{c_1, c_2, ..., c_k\}$, where $k$ is the total number of object categories. We define the context graph (an example shown in Figure~\ref{fig:intro} b) as a fully connected directed graph, in which each node is associated with an object category $c_i$ and the weight on the edge $e_{i,j}$ encodes different properties relating two nodes such as their co-occurrence probability, distance, and relative size. The number of nodes in the context graph is same as the number of object categories $k$.

\paragraph*{Co-occurrence graph.}
We aim to model the co-occurrence probability of each pair of instances in a natural image. To be more specific, we seek to determine the probability of the event that an instance of category $c_j$ appears in the image {\em given that} an instance of category $c_i$ also appears in the image.
Co-occurrence graph inherits the structure of the aforementioned context graph, and the directed edge $e_{i,j}$ represents the probability that an instance of category $c_j$ appears in the image given an instance $c_i$ already exists. This probability is denoted by $p^\text{occur}_{i,j}=p(c_j|c_i)$. Note that for each node, we also have an edge pointing to itself (i.e., $e_{i,i}$). According to the definition of probability we have $0 \leq p^\text{occur}_{i,j} \leq 1$ and $\sum_{j=1,\ldots,k}{p^\text{occur}_{i,j} = 1 }$ for all $i$.  

To compose such a co-occurrence graph, we can calculate a matrix $P=\{p^\text{occur}_{i,j} | i,j=1,\ldots,k\}$ using a large-scale natural scene image dataset $\mathcal{D'}$, whose distribution is deemed to be similar to  $\mathcal{D}$. We approximate co-occurrence probabilities using the relative co-occurrence frequencies of objects from $\mathcal{D'}$.

\paragraph*{Distance graph.}
Suppose the bounding box of an object is given as $[x^c, y^c, h, w]$, where $(x^c,y^c)$ denote the center pixel location of the box and $(h, w)$ denote the height and width in pixels. 
%
In distance graph, the edge $e_{i,j}$ captures the distribution of the $\ell_2$ distance between center points of $c_j$ and $c_i$. The bilateral edges are equivalent. Considering the fact that the image size $(H,W)$ varies in the dataset, which will also influence the distance between two objects, thus to minimize this scaling effect, we normalize the distance by image diagonal $L = \sqrt{H^2+W^2}$. The distance distribution is denoted as $p^\text{dist}_{i,j}(\ell_2 ([x_{i}^{c},y_{i}^{c}],[x_{j}^{c},y_{j}^{c}]) / L | c_i)$.

\paragraph*{Size graph.}
Similarly, size graph models the 2D distributions of object height and width, where edge $e_{i,j}$ represents $p^\text{size}_{i,j}(h_j / L, w_j / L | c_i)$, which is the distribution of height and width of $c_j$ given $c_i$ is also present in the image.

\subsection{Context-Aware Attack Plan} 
Given an image $I$, we denote the instance categories in the image as $X=[x_1, x_2, ..., x_m]$, where $m$ is the total number of detected objects in the image. Note that different $x_i$ could be the same because two instances of the same category can co-occur in a scene. 

For the miscategorization attack, the goal is to miscategorize $x_i$ to $x'_{i}$ for   $i\in\{1,\ldots,m\}$. We call the object associated with $x_i$ as the \textbf{victim object} or the \textbf{victim instance}. %
To simplify our discussion, let us assume that our goal is to miscategorize $x_1$ to $x'_1$. 
Note here that methods that focus on miscategorizing the victim object/instance only will search for a perturbation so that the labels for all the objects become $X'=[x'_1, x_2, ..., x_m]$. We call $X'$ the {\bf  attack plan}, since it yields the target labels for attackers. 

In our proposed context-aware attack method, in addition to miscategorizing $x_1$ into $x'_1$, we may also want to miscategorize one or more helper objects that can provide important context information for $x_1$. We create a context-aware attack plan as $X'_c =[x'_1, x'_2, ..., x'_n]$. We will use subscript $c$ with context-aware attack plans to distinguish them from the context-agnostic attack plans.
%
The $x'_i$ could be the same as $x_i$ when we do not seek to miscategorize the instance associated with $x_i$. 
All the $x_i$ (except the victim object) that change to a different label $x'_i$ in $X'_c$ are called  \textbf{helper instances}.  
Note that in the context-aware attack plan, $X'_c$, $n$ could be greater than $m$ in cases where we decide to insert new instances as helper objects. 
We illustrate an example attack plan in Figure~\ref{fig:intro}(c), where the bird at the bottom is the victim object that we want to mis-categorize to a table; the bird at the top (to be mis-categorized as chair) and the additional appearing chair are the two helper instances. 

The number of helper instances is a hyper-parameter that we tune. We use the co-occurrence graph, defined previously, to decide which existing instances $(x_i)$ should serve as helper instances, and what category labels $(x'_i)$ should be assigned to these instances. 
From the co-occurrence graph, we obtain the co-occurrence probability with respect to every possible instance pair category. 
Given the goal of miscategorizing victim object from $x_1$ to $x'_1$, we choose the label for every helper instance $(x'_i)$ by sampling the label space $C$ according to the co-occurrence probability $p(x'_i = c | x'_1)$ for all $c\in C$. Note that $\sum_{c\in C}p(x_i' = c|x'_1)=1$.  
%
We could model the joint probability of all helper instances given the target label, but that would require a large amount of data. Our sampling approach assumes conditional independence of helper instances (akin to na\"ive Bayes), in which we draw the most probable labels for our helper labels by sampling one row of the co-occurrence probability matrix.
By random sampling the label space in this manner, we expect that objects that occur more frequently will be selected as labels for the helper objects.
We first select the helper objects from among the $m$ objects present in the scene. In case we need to add new helper instances ($>m$), we choose their locations and sizes according to the mean values of the distributions given by distance and size graphs. %

\subsection{Sequential Attack Generation}\label{sec:sequential}

We propose a sequential perturbation generation strategy, where we start with zero helper objects in the attack plan and sequentially add one helper object until the attack succeeds on the blackbox, as shown in Figure~\ref{fig:intro}.
We generate adversarial attacks using a single or multiple surrogate model(s) in our perturbation machine. As we sequentially add the helper objects in the attack plan, we query the black-box model to see if our attack succeeds.  In our experiments, we make up to 6 queries to the blackbox detector, which provides hard labels for the detected objects. We use this information only as a stopping criterion for the attack generation. We stop the sequential attack process if the adversarial example fools the black-box model or we run out of the budget of helper objects. 
Note that our strategy is orthogonal to query-based methods that aim to generate adversarial examples or estimate gradients of the blackbox models (often using hundreds or thousands of queries) \cite{Wang2020blackobject, Cheng2019Improving, huang2019black}.

Our attack generation method with a single surrogate detector is based on targeted adversarial objectness gradient attacks (TOG) \cite{chow2020adversarial}, which can be viewed as training the detector for modified labels given in the attack plan $X'$. The weights of the detector network remain fixed but a perturbation image $\delta$ is added to the clean image as $I + \delta$ at every iteration to minimize the training loss $\mathcal{L}(\text{clip}(I+\delta);\mathcal{O'})$ for a desired output $\mathcal{O'}$. The value of $I + \delta$ is clipped at each iteration to make sure it is legally bounded. We generate the desired output $\mathcal{O'}$ based on our attack plan $X'$. The attack plan in $X'$ only contains label information, but we also assign location and confidence score information in $\mathcal{O'}$. At every iteration, we update the perturbation using the iterative fast gradient signed method (I-FGSM), 
\begin{equation}\label{eq:IFGSM}
\delta \leftarrow \delta - \epsilon \cdot \text{sign}[\nabla_{\delta}\mathcal{L}(\text{clip}(I+\delta);\mathcal{O'})],
\end{equation} 
where $\epsilon$ is the step size at each iteration. We can also use an ensemble of detectors as the surrogate models in perturbation machine, where we generate perturbation by minimizing the joint loss function over all detectors:
\begin{equation}\label{eq:ensemble_loss}
\mathcal{L} = \alpha_{1} \mathcal{L}_{\text{1}} + \alpha_{2} \mathcal{L}_{\text{2}} + ... + \alpha_{N} \mathcal{L}_{\text{N}},
\end{equation} 
while keeping $\sum_{i=1}^{N} \alpha_i = 1$ and $\alpha_i >0$ for all $i$. 
%
%
We can easily modify our method to use other perturbation generation methods and loss functions  \cite{madry2017towards,carlini2017towards,dong2018boosting, xie2019improving,lin2019nesterov, wang2021enhancing}.
\begin{table*}[!ht]
\centering
\caption{White-box and black-box mis-categorization attack fooling rate on different models with different perturbation budgets ($L_{\infty} \leq \{10,20,30\}$) using VOC and COCO dataset. Baseline only perturbs the victim object, while ours also perturbs other objects conforming to context. Random perturbs other objects but assign random labels. \textbf{Abbreviation:} Faster R-CNN (FRCNN), RetinaNet (Retina), Libra R-CNN (Libra), FoveaBox (Fovea), FreeAnchor (Free), Deformable DETR (D-DETR).
}
\label{tab:combined}
\small
\begin{tabular}{cccc|cccccc}
\toprule 
\multirow{2}{*}{\begin{tabular}[c]{@{}c@{}}Perturbation \\ Budget\end{tabular}} &
  \multirow{2}{*}{Method} &
  \multicolumn{2}{c|}{\textbf{Whitebox}} &
  \multicolumn{6}{c}{\textbf{Blackbox}} \\ 
                                   &          & FRCNN & YOLOv3 & Retina & Libra & Fovea & Free & DETR & D-DETR \\ \hline

\addlinespace
\multicolumn{10}{c}{\textbf{\textit{Results on PASCAL VOC}}}\\

\multirow{3}{*}{$L_{\infty} \le 10$} & Baseline & 40.0  & 53.8   & 13.8   & 9.2   & 22.2  & 27.4 & 9.6  & 23.2   \\
                                   & Random & 52.4 & 69.2 & 19.4 & 17.4 & 31.6 & 37.8 & 17.4 & 36.8 \\
                                    & Ours & \textbf{55.8} & \textbf{75.6} & \textbf{22.6} & \textbf{20.4} & \textbf{33.6} & \textbf{39.2} & \textbf{20.2} & \textbf{39.2} \\ \hline
\multirow{3}{*}{$L_{\infty} \le 20$} & Baseline & 65.2  & 67.8   & 24.0   & 21.4  & 34.4  & 41.8 & 14.4 & 37.6   \\
                                   & Random & 74.4 & 83.8 & 31.0 & 29.6 & 46.2 & 54.4 & 28.0 & 52.6 \\
                                    & Ours & \textbf{78.6} & \textbf{87.2} & \textbf{35.2} & \textbf{38.4} & \textbf{51.6} & \textbf{56.6} & \textbf{34.0} & \textbf{58.4} \\ \hline
\multirow{3}{*}{$L_{\infty} \le 30$} & Baseline & 70.6  & 70.4   & 29.8   & 28.6  & 41.6  & 48.0 & 20.4 & 38.6   \\
                                    & Random & 79.2 & 82.6 & 37.8 & 36.8 & 53.4 & 59.8 & 34.4 & 52.8 \\
                                    & Ours & \textbf{80.6} & \textbf{88.0} & \textbf{42.0} & \textbf{44.2} & \textbf{56.8} & \textbf{63.6} & \textbf{40.2} & \textbf{59.0} \\ \hline

\addlinespace
\multicolumn{10}{c}{\textbf{\textit{Results on MS COCO}}}\\

\multirow{3}{*}{$L_{\infty} \le 10$} & Baseline & 29.0  & 32.2  & 7.4    & 4.8   & 11.6  & 16.6 & 3.4  & 19.0   \\
                                   & Random & 40.2 & 48.4 & 11.2 & 8.0 & 14.6 & 20.0 & 6.2 & 23.6 \\
 & Ours & \textbf{41.2} & \textbf{54.4} & \textbf{12.0} & \textbf{11.2} & \textbf{18.6} & \textbf{25.0} & \textbf{10.8} & \textbf{27.8} \\ \hline
\multirow{3}{*}{$L_{\infty} \le 20$} & Baseline & 51.8  & 49.2  & 13.4   & 11.8  & 22.0  & 28.6 & 8.8  & 26.8   \\
                                   & Random & 60.6 & 66.4 & 20.6 & 18.8 & 31.4 & 37.2 & \textbf{20.2} & 39.2 \\
 & Ours & \textbf{64.4} & \textbf{70.0} & \textbf{20.8} & \textbf{22.2} & \textbf{35.4} & \textbf{40.8} & 20.0 & \textbf{43.2} \\ \hline
\multirow{3}{*}{$L_{\infty} \le 30$} & Baseline & 57.6  & 54.4  & 18.2   & 15.4  & 25.6  & 34.8 & 8.0  & 28.8   \\
                                   & Random & 65.8 & 73.6 & 23.8 & 21.8 & 34.8 & \textbf{47.8} & 18.4 & 42.0 \\
 & Ours & \textbf{68.6} & \textbf{75.4} & \textbf{27.2} & \textbf{27.2} & \textbf{39.2} & 46.2 & \textbf{21.2} & \textbf{48.6} \\
\bottomrule
\end{tabular}
\end{table*}

\section{Experiments}
We perform comprehensive experiments on two large-scale object detection datasets to evaluate the proposed context-aware sequential attack strategy.
We mainly show that the context-aware sequential attack strategy can help with mis-categorization attacks in blackbox setting. We also present results with whitebox setting, for completeness, even though this is not our primary objective.
%

\subsection{Implementation Details}
\PP{Object detection models.} We evaluate our attack plans on a diverse set of object detectors, including 
\begin{itemize} 
    \item \textbf{Two-stage detectors.} Faster R-CNN~\cite{ren2015faster}, Libra R-CNN~\cite{pang2019libra,pang2021towards};
    \item \textbf{One-stage detectors.} YOLOv3~\cite{redmon2018yolov3}, RetinaNet~\cite{lin2017focal};
    \item \textbf{Anchor-free detectors.} FoveaBox~\cite{kong2020foveabox}, FreeAnchor~\cite{zhang2019freeanchor};
    \item \textbf{Transformer-based detectors.} DETR~\cite{carion2020end}, Deformable DETR~\cite{zhu2021deformable}.
\end{itemize}

We use \texttt{MMDetection} \cite{mmdetection} code repository for the aforementioned models. Inspired by ~\cite{Liu2017Delving,wu2020making}, we use an ensemble of locally trained object detection models as the surrogate model. 
%
Selecting a good surrogate ensemble is an interesting question, where the number and type of surrogate models will influence the attack success rate. We tested different single and multiple models as surrogates in our preliminary tests and observed a similar trend that the context-aware attacks significantly outperform the baseline attacks that are context-agnostic. Therefore, we selected two most commonly-used models, Faster R-CNN and YOLOv3, as the surrogate ensemble in our experiments. The weighting factor $\alpha$ is chosen such that the individual loss terms are balanced. On the blackbox victim side, we choose the leftover models that have a variety of different architectures.

\PP{Datasets.} 
We use images from both PASCAL VOC~\cite{everingham2010pascal} and MS COCO~\cite{lin2014microsoft} datasets in our experiments. VOC contains 20 object categories which commonly appear in natural environment, and COCO contains 80 categories which is a super-set of the categories in VOC. We randomly selected 500 images that contain multiple $(2-6)$ objects from \texttt{voc2007test} and \texttt{coco2017val}. Since all models in \texttt{MMDetection} are trained on \texttt{coco2017train}, while testing the detectors on VOC images, we only return the objects that also exist in VOC categories.

\PP{Context graph construction.} For VOC and COCO images, we extract context from \texttt{voc2007trainval} and \texttt{coco2017train} respectively. For each dataset, we build three $N\times N$ arrays ($N$ is number of labels) that contain co-occurrence probability, distance distribution, and size distributions. The $(i,j)$ cell in the co-occurrence array stores the number of co-occurrences of object $c_i$ and object $c_j$ normalized by the summation of that row; each cell in the distance array is a 1D distribution of the distances between $c_i$ and $c_j$ found in the images; each cell in the size table is a 2D distribution of $h$ and $w$ of $c_j$ given $c_i$. These three arrays can be easily computed form the datasets.

\PP{Attack generation.}
We use I-FGSM-based method to generate a perturbation on the whole image (as discussed in Eqn. \eqref{eq:IFGSM}), and we limit the maximum perturbation level to be $L_{\infty} \leq \{10,20,30\}$. The number of helper objects is empirically chosen to be $5$. We present an analysis study on how the attack performance changes with the number of helper objects in Section~\ref{subsec:analysis} of analysis study.

\PP{Baseline and comparisons.} TOG \cite{chow2020adversarial} shows better performance compared to UEA \cite{Wei2019Transferable} and RAP \cite{li2018robust}; therefore, to understand the performance of the proposed context-aware attack plan strategy, we use the current state-the-art attack strategy based on TOG \cite{chow2020adversarial}. The attack plan generated by the baseline (labeled as Baseline in Table~\ref{tab:combined}) is context-agnostic and only associated with the victim object. To validate that our proposed context-aware attack really benefits from co-occurrence, location and size information, we also present results for a setting (labeled as Random in Table~\ref{tab:combined}) in which we choose helper objects label and location at random.

\begin{figure*}[!ht]
\centering 
\hspace*{-1cm} 
\includegraphics[scale=0.7]{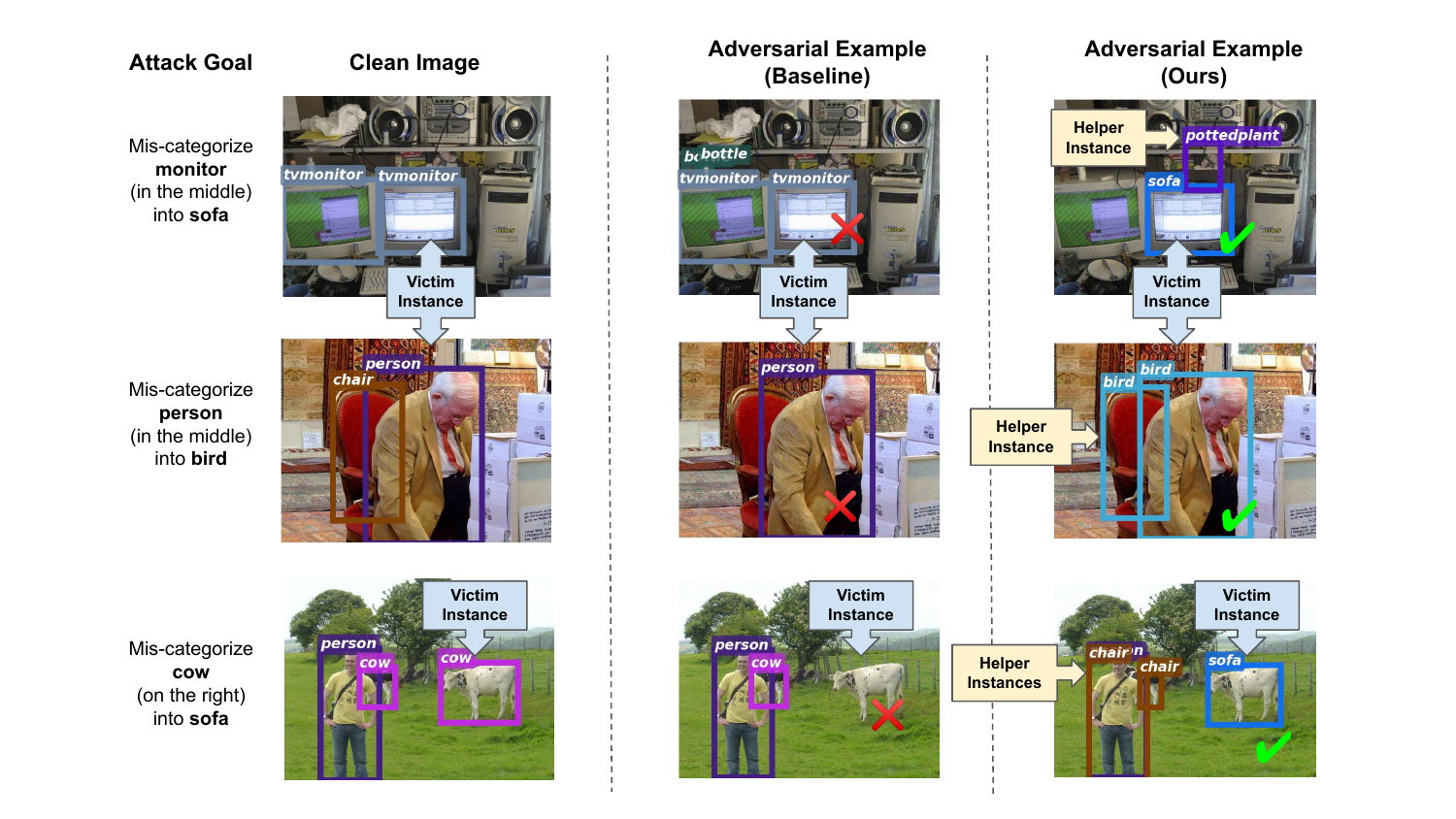}
\caption{Examples where baseline attack fails but context-aware method succeeds by introducing helper objects in the attack. The perturbation ($L_{\infty} \le 10$) is generated from our perturbation machine (whitebox ensemble of FRCNN and YOLOv3) and tested on the blackbox model (RetinaNet). The detection results on original image, image perturbed by baseline attack, and image perturbed by our context-aware method are shown in the subfigures from left to right. In these examples, we introduce \texttt{pottedplant} as a helper object to mis-categorize the victim \texttt{monitor} to \texttt{sofa}, introduce another \texttt{bird} to mis-categorize the \texttt{person} to a \texttt{bird}, and add a few \texttt{chairs} to mis-categorize the \texttt{cow} to a \texttt{sofa}. Visualization of perturbation level $L_{\infty} <= 20, 30$ can be found in supplementary materials.
}
\label{fig:visualization}
\end{figure*}

\PP{Evaluation metric} 
We use attack success rate (or fooling rate) to evaluate the adversarial attack performance on any victim object detector. Since we perform targeted mis-categorization attack, instead of using mAP which takes account of all existing objects, we only focus on the victim object and define our attack success rate as the percentage of attacks in which the victim object was successfully mis-classified to the target label. In experiments, we check if the target object exists in the detection with an intersection over union (IOU) greater than $0.3$. If yes, the attack is successful (or the detector is fooled); otherwise, the attack fails. For the selection of target objects, we randomly selected one target label that is not present in the original image to mimic the out-of-context attack as well as eliminating the chance of miscounting the existing objects as success. 

\begin{figure*}[!ht]
    \centering 
    \begin{subfigure}[t]{0.4\linewidth}
    \includegraphics[width=1\linewidth]{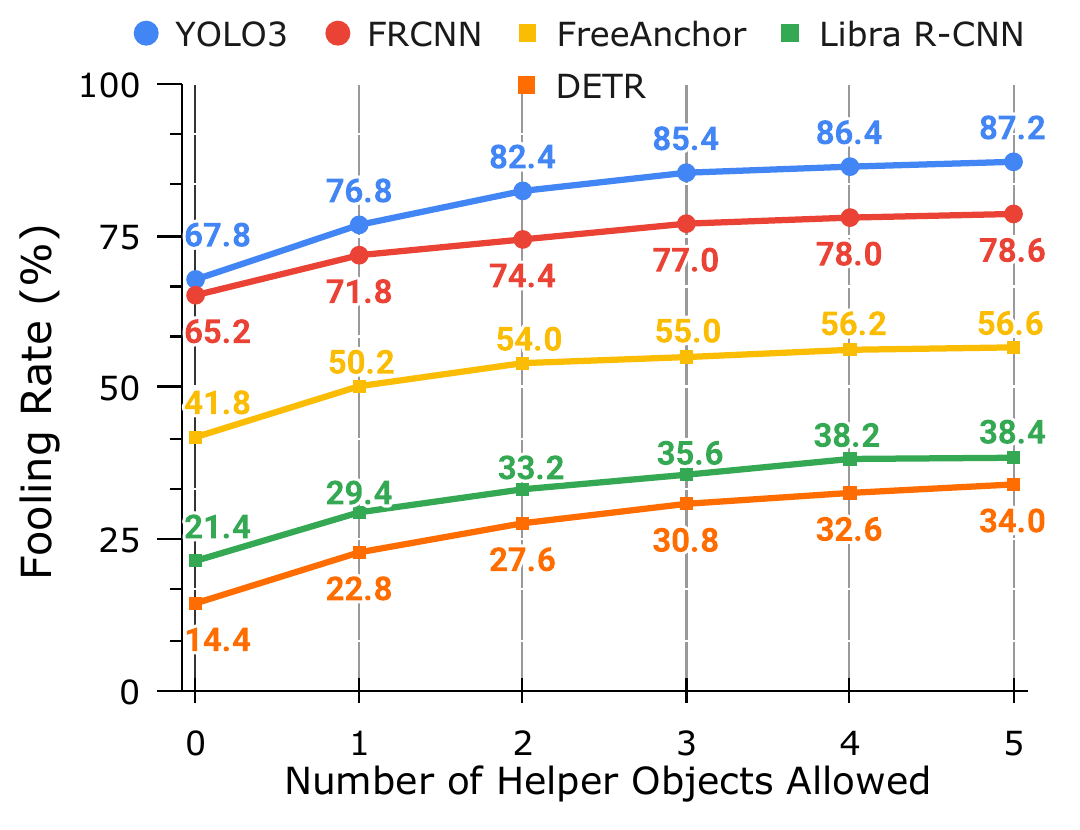}
    \caption{PASCAL VOC, $L_{\infty} \le 20$}
    \end{subfigure}
    ~~
    \begin{subfigure}[t]{0.4\linewidth}
    \includegraphics[width=1\textwidth]{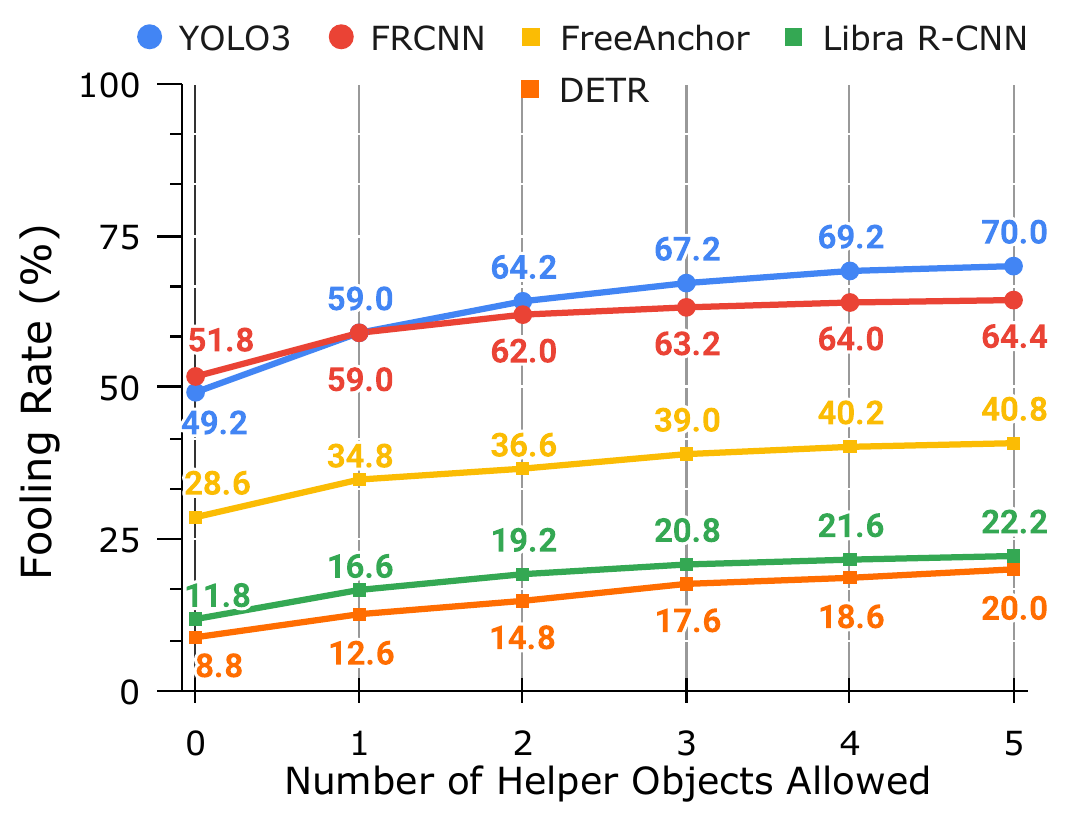}
    \caption{MS COCO, $L_{\infty} \le 20$}
    \end{subfigure}
    \caption{Mis-categorization attack fooling rate of white-box and black-box models at perturbation level $L_{\infty} \le 20$ w.r.t. number of helper objects allowed (changed or added). Circles denote white-box models (FRCNN and YOLO3) and squares denote black-box models (FreeAnchor, Libra R-CNN, and DETR). 
    Plots of perturbation level $L_{\infty} \le 10, 30$ can be found in supplementary material.}
    \label{fig:analysis_study_mis}
\end{figure*}

\subsection{Evaluation of Attack Performance}
\PP{Whitebox attack performance.}
We observe that the attack success rate suffers even in whitebox setting, especially when the perturbation budget is small. As shown in Table~\ref{tab:combined}, the baseline whitebox attack with $L_{\infty} \le 10$ on COCO can only achieve around $30\%$ fooling rate. This is because we simultaneously attack multiple objects in the image and also use an ensemble loss to fool multiple models jointly, the targeted mis-categorization attack is challenging. Even in this difficult setting, our context-aware attack can successfully improve the fooling rate by $10-20$ percentage points. Besides this, we can observe that our method provides significant improvement (by at least 10 percentage points) over the baseline method at all perturbation levels on both VOC and COCO dataset. Our performance is not only better than baseline method, but also has clear advantage over sequential attacks with random context. This validates the effectiveness of the proposed context-aware sequential attack strategy in the whitebox settings.

\PP{Blackbox attack performance.}
We test the performance of the attacks generated by the surrogate detectors in the perturbation machine on different blackbox detectors. Our hypothesis is that the context-aware adversarial examples transfer better to the unseen models, and thus have better attack performance compared to context-agnostic (baseline) attacks in the blackbox setting. We use the same baseline and evaluation metrics as in the evaluation of the whitebox attack. Our results corroborate our hypothesis as we observe that even though the blackbox attack success rate is significantly lower compared to the whitebox attack success rate, our proposed context-aware sequential attack strategy still provides significantly better transfer success rate compared to the context-agnostic (baseline) attacks. For both VOC and COCO datasets, for all levels of perturbation. Overall, for every test setting, our method improves the success rate over baseline method by 5--20 percentage points (average improvement is beyond 10 percentage points).
This is a significant improvement for the notoriously difficult problem of transfer attacks on object detectors in blackbox settings by using just 2--6 queries. 
Our proposed context-aware attack strategy has better transfer rates than the context-agnostic baseline and random assignment of labels, which further shows the benefits of utilizing co-occurrence relationships, location and size information to generate the attack plans.

\PP{Visualization.}
We show three attack examples in Figure~\ref{fig:visualization}. In the first example, we aim to miscategorize a TV monitor to sofa. We observe that the baseline attack fails to transfer to the blackbox model, RetinaNet (middle row). In comparison, the context-aware adversarial example from our method fools the victim blackbox model to detect the TV monitor as a sofa by introducing a pottedplant as the helper object, which frequently co-occurs with the target label, sofa. In the second example, we aim to miscategorize a person into a bird. The baseline attack fails since the person is still detected. However, our method succeeds by introducing another bird as the helper object. In the third example, we seek to mis-categorize a cow as a sofa. The baseline attack fails as no object is detected near the victim object. Our context-aware attack plan succeeds by assigning the person and other cow in the image to chairs (helper objects). 

 
\subsection{Analysis Study}
\label{subsec:analysis}
\PP{Number of helper objects.} \label{analysis:number-helpers}
Even though helper objects boost the adversarial attack success, we do not need a large number of them. Since the perturbation budget is fixed, using too many helper instances may reduce the effect for the victim  instance. On the other hand, not using any helper instances would completely eliminate the benefits of context-aware attacks. 
%
To investigate how the number of helper objects affects the attack performance, we plot the mis-categorization attack success rate with respect to the number of helper objects in Figure~\ref{fig:analysis_study_mis}. 
 We observe that adding more objects improves attack success rate both for the whitebox and blackbox attacks. The improvement is profound for some blackbox attacks that have low baseline attack success rates. We also observe that the first few helper objects boost the attack performance significantly and the improvement gradually plateaus as we add 4--5 helper objects.

\begin{figure}[t]
\centering 
\includegraphics[width=0.85\linewidth,trim=3.5cm 0.9cm 2.8cm 0.9cm, clip]{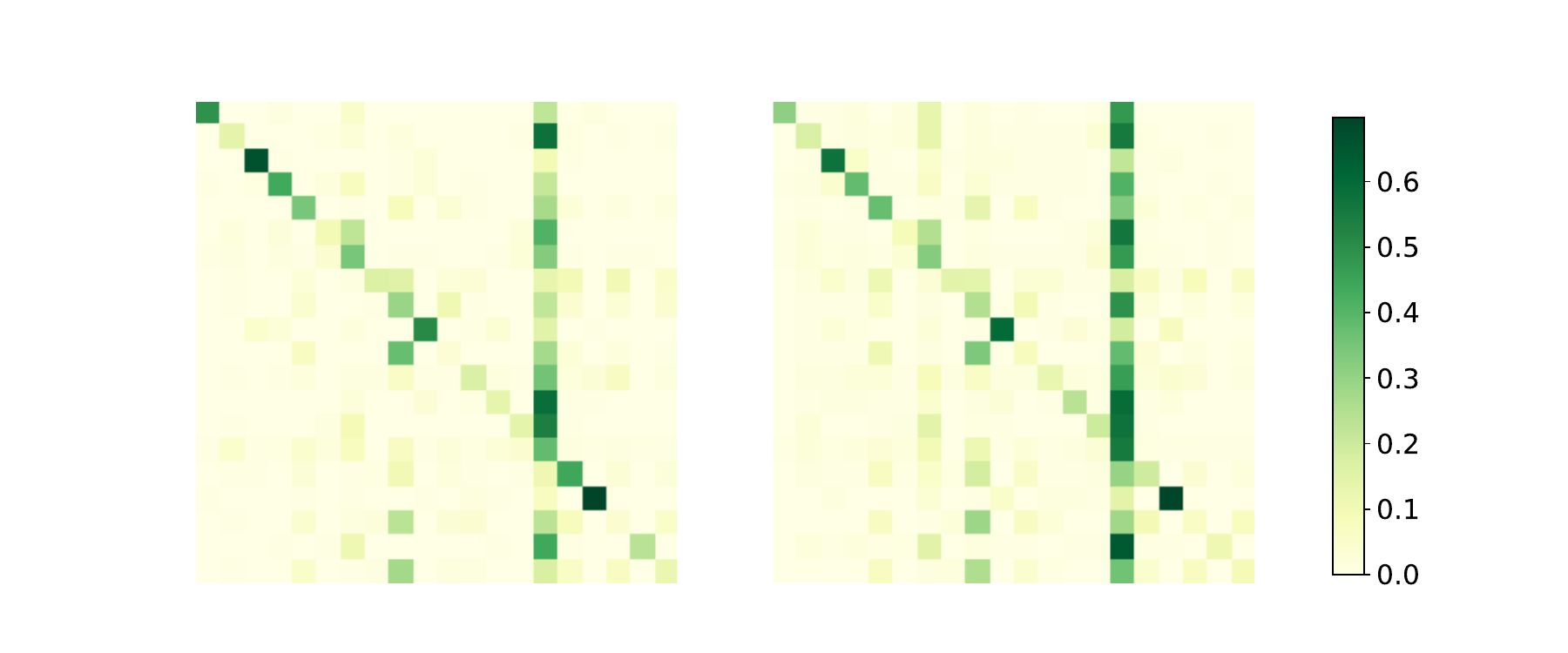}
\caption{Co-occurrence matrices for VOC (left) and COCO (right) for 20 object categories that are common in both datasets.
}
\label{fig:co-mat}
\end{figure}

\PP{Context graphs of different datasets.} To demonstrate that the context graphs are generic enough to be used across different natural scene datasets, we evaluate the similarity of the co-occurrence matrices extracted from the two large-scale datasets (VOC and COCO). The average Perason correlation coefficient of each corresponding row of VOC matrix and COCO matrix is $0.90$, which signifies strong positive correlation between co-occurrence relationships encoded by these two context graphs. We can visually see the similarities of these two co-occurrence matrices in Figure~\ref{fig:co-mat}. One of the salient patterns common in these two matrices is that the column of \texttt{person} is colored in dark green, showing that person generally has a high probability to co-occur with other objects. This is a notable feature of natural scene images.
Because of the high similarity of the contexts in the two datasets, we can use their context graphs interchangeably. It is indeed possible that if the original context of objects in the given image is very different from the context graph we use to build the attack plan, the transfer attack success rate will suffer. This can be corroborated by the comparison of Random and Ours in Table~\ref{tab:combined}.

\section{Conclusion}
In this paper, we propose a novel context-aware adversarial attack method that exploits rich object co-occurrence relationships plus location and size information to effectively improve mis-categorization attack fooling rate against blackbox object detectors. Our experimental results on two large-scale datasets show that our attack success rate is significantly higher than baseline and comparing methods, which validates the effectiveness of our methods. The contextual relationships modeled by our method holds true in different datasets within natural image domain, thus implying the wide applicability of our methods. 

\section{Acknowledgments.} 
This material is based upon work supported by the Defense Advanced Research Projects Agency (DARPA) under agreement number HR00112090096.
Approved for public release; distribution is unlimited.

\bibliography{aaai22.bib}

\begin{thebibliography}{55}
\providecommand{\natexlab}[1]{#1}

\bibitem[{Barnea and Ben-Shahar(2019)}]{barnea2019exploring}
Barnea, E.; and Ben-Shahar, O. 2019.
\newblock Exploring the bounds of the utility of context for object detection.
\newblock In \emph{IEEE Conference on Computer Vision and Pattern Recognition},
  7412--7420.

\bibitem[{Bell et~al.(2016)Bell, Zitnick, Bala, and Girshick}]{bell2016inside}
Bell, S.; Zitnick, C.~L.; Bala, K.; and Girshick, R. 2016.
\newblock Inside-outside net: Detecting objects in context with skip pooling
  and recurrent neural networks.
\newblock In \emph{IEEE Conference on Computer Vision and Pattern Recognition},
  2874--2883.

\bibitem[{Brendel, Rauber, and Bethge(2018)}]{brendel2017decision}
Brendel, W.; Rauber, J.; and Bethge, M. 2018.
\newblock Decision-based adversarial attacks: Reliable attacks against
  black-box machine learning models.
\newblock In \emph{International Conference on Learning Representations}.

\bibitem[{Carion et~al.(2020)Carion, Massa, Synnaeve, Usunier, Kirillov, and
  Zagoruyko}]{carion2020end}
Carion, N.; Massa, F.; Synnaeve, G.; Usunier, N.; Kirillov, A.; and Zagoruyko,
  S. 2020.
\newblock End-to-end object detection with transformers.
\newblock In \emph{European Conference on Computer Vision}, 213--229.

\bibitem[{Carlini and Wagner(2017)}]{carlini2017towards}
Carlini, N.; and Wagner, D. 2017.
\newblock Towards evaluating the robustness of neural networks.
\newblock In \emph{IEEE Symposium on Security and Privacy}, 39--57.

\bibitem[{Chen, Jordan, and Wainwright(2020)}]{chen2020hopskipjumpattack}
Chen, J.; Jordan, M.~I.; and Wainwright, M.~J. 2020.
\newblock Hopskipjumpattack: A query-efficient decision-based attack.
\newblock In \emph{IEEE Symposium on Security and Privacy}, 1277--1294.

\bibitem[{Chen et~al.(2019)Chen, Wang, Pang, Cao, Xiong, Li, Sun, Feng, Liu,
  Xu, Zhang, Cheng, Zhu, Cheng, Zhao, Li, Lu, Zhu, Wu, Dai, Wang, Shi, Ouyang,
  Loy, and Lin}]{mmdetection}
Chen, K.; Wang, J.; Pang, J.; Cao, Y.; Xiong, Y.; Li, X.; Sun, S.; Feng, W.;
  Liu, Z.; Xu, J.; Zhang, Z.; Cheng, D.; Zhu, C.; Cheng, T.; Zhao, Q.; Li, B.;
  Lu, X.; Zhu, R.; Wu, Y.; Dai, J.; Wang, J.; Shi, J.; Ouyang, W.; Loy, C.~C.;
  and Lin, D. 2019.
\newblock {MMDetection}: Open MMLab Detection Toolbox and Benchmark.
\newblock \emph{arXiv preprint arXiv:1906.07155}.

\bibitem[{Chen et~al.(2017)Chen, Zhang, Sharma, Yi, and Hsieh}]{Chen2017zoo}
Chen, P.-Y.; Zhang, H.; Sharma, Y.; Yi, J.; and Hsieh, C.-J. 2017.
\newblock Zoo: Zeroth order optimization based black-box attacks to deep neural
  networks without training substitute models.
\newblock In \emph{Proceedings of the 10th ACM workshop on artificial
  intelligence and security}, 15--26.

\bibitem[{Chen, Huang, and Tao(2018)}]{chen2018context}
Chen, Z.; Huang, S.; and Tao, D. 2018.
\newblock Context refinement for object detection.
\newblock In \emph{European Conference on Computer Vision}, 71--86.

\bibitem[{Cheng et~al.(2019)Cheng, Dong, Pang, Su, and
  Zhu}]{Cheng2019Improving}
Cheng, S.; Dong, Y.; Pang, T.; Su, H.; and Zhu, J. 2019.
\newblock {Improving black-box adversarial attacks with a transfer-based
  prior}.
\newblock \emph{Advances in Neural Information Processing Systems},
  10934--10944.

\bibitem[{Chow et~al.(2020)Chow, Liu, Loper, Bae, Gursoy, Truex, Wei, and
  Wu}]{chow2020adversarial}
Chow, K.-H.; Liu, L.; Loper, M.; Bae, J.; Gursoy, M.~E.; Truex, S.; Wei, W.;
  and Wu, Y. 2020.
\newblock Adversarial Objectness Gradient Attacks in Real-time Object Detection
  Systems.
\newblock In \emph{2020 Second IEEE International Conference on Trust, Privacy
  and Security in Intelligent Systems and Applications (TPS-ISA)}, 263--272.

\bibitem[{Deng et~al.(2009)Deng, Dong, Socher, Li, Li, and
  Fei-Fei}]{deng2009imagenet}
Deng, J.; Dong, W.; Socher, R.; Li, L.-J.; Li, K.; and Fei-Fei, L. 2009.
\newblock Imagenet: A large-scale hierarchical image database.
\newblock In \emph{IEEE Conference on Computer Vision and Pattern Recognition},
  248--255.

\bibitem[{Divvala et~al.(2009)Divvala, Hoiem, Hays, Efros, and
  Hebert}]{divvala2009empirical}
Divvala, S.~K.; Hoiem, D.; Hays, J.~H.; Efros, A.~A.; and Hebert, M. 2009.
\newblock An empirical study of context in object detection.
\newblock In \emph{IEEE Conference on Computer Vision and Pattern Recognition},
  1271--1278.

\bibitem[{Dong et~al.(2018)Dong, Liao, Pang, Su, Zhu, Hu, and
  Li}]{dong2018boosting}
Dong, Y.; Liao, F.; Pang, T.; Su, H.; Zhu, J.; Hu, X.; and Li, J. 2018.
\newblock {Boosting Adversarial Attacks with Momentum}.
\newblock In \emph{IEEE Conference on Computer Vision and Pattern Recognition},
  9185--9193.

\bibitem[{Everingham et~al.(2010)Everingham, Van~Gool, Williams, Winn, and
  Zisserman}]{everingham2010pascal}
Everingham, M.; Van~Gool, L.; Williams, C.~K.; Winn, J.; and Zisserman, A.
  2010.
\newblock The pascal visual object classes (voc) challenge.
\newblock \emph{International Journal of Computer Vision}, 88(2): 303--338.

\bibitem[{Galleguillos and Belongie(2010)}]{galleguillos2010context}
Galleguillos, C.; and Belongie, S. 2010.
\newblock Context based object categorization: A critical survey.
\newblock \emph{Computer Vision and Image Understanding}, 114(6): 712--722.

\bibitem[{Goodfellow, Shlens, and Szegedy(2015)}]{goodfellow2014explaining}
Goodfellow, I.~J.; Shlens, J.; and Szegedy, C. 2015.
\newblock Explaining and harnessing adversarial examples.
\newblock \emph{International Conference on Learning Representations}.

\bibitem[{Guo et~al.(2019)Guo, Gardner, You, Wilson, and
  Weinberger}]{guo2019simple}
Guo, C.; Gardner, J.; You, Y.; Wilson, A.~G.; and Weinberger, K. 2019.
\newblock Simple black-box adversarial attacks.
\newblock In \emph{International Conference on Machine Learning}, 2484--2493.

\bibitem[{Huang and Zhang(2020)}]{huang2019black}
Huang, Z.; and Zhang, T. 2020.
\newblock Black-box adversarial attack with transferable model-based embedding.
\newblock \emph{International Conference on Learning Representations}.

\bibitem[{Kong et~al.(2020)Kong, Sun, Liu, Jiang, Li, and
  Shi}]{kong2020foveabox}
Kong, T.; Sun, F.; Liu, H.; Jiang, Y.; Li, L.; and Shi, J. 2020.
\newblock Foveabox: Beyound anchor-based object detection.
\newblock \emph{IEEE Transactions on Image Processing}, 29: 7389--7398.

\bibitem[{Li et~al.(2020{\natexlab{a}})Li, Xu, Zhang, Yang, and
  Li}]{li2020qeba}
Li, H.; Xu, X.; Zhang, X.; Yang, S.; and Li, B. 2020{\natexlab{a}}.
\newblock Qeba: Query-efficient boundary-based blackbox attack.
\newblock In \emph{IEEE Conference on Computer Vision and Pattern Recognition},
  1221--1230.

\bibitem[{Li et~al.(2020{\natexlab{b}})Li, Deng, Li, Yan, Gao, and
  Huang}]{li2020towards}
Li, M.; Deng, C.; Li, T.; Yan, J.; Gao, X.; and Huang, H. 2020{\natexlab{b}}.
\newblock Towards transferable targeted attack.
\newblock In \emph{IEEE Conference on Computer Vision and Pattern Recognition},
  641--649.

\bibitem[{Li et~al.(2020{\natexlab{c}})Li, Zhu, Paul, Roy-Chowdhury, Song,
  Krishnamurthy, Swami, and Chan}]{li2020connecting}
Li, S.; Zhu, S.; Paul, S.; Roy-Chowdhury, A.; Song, C.; Krishnamurthy, S.;
  Swami, A.; and Chan, K.~S. 2020{\natexlab{c}}.
\newblock Connecting the dots: Detecting adversarial perturbations using
  context inconsistency.
\newblock In \emph{European Conference on Computer Vision}, 396--413.

\bibitem[{Li et~al.(2018)Li, Tian, Chang, Bian, and Lyu}]{li2018robust}
Li, Y.; Tian, D.; Chang, M.-C.; Bian, X.; and Lyu, S. 2018.
\newblock Robust adversarial perturbation on deep proposal-based models.
\newblock \emph{The British Machine Vision Conference}, 231.

\bibitem[{Lin et~al.(2020)Lin, Song, He, Wang, and Hopcroft}]{lin2019nesterov}
Lin, J.; Song, C.; He, K.; Wang, L.; and Hopcroft, J.~E. 2020.
\newblock Nesterov accelerated gradient and scale invariance for adversarial
  attacks.
\newblock In \emph{International Conference on Learning Representations}.

\bibitem[{Lin et~al.(2017)Lin, Goyal, Girshick, He, and
  Doll{\'a}r}]{lin2017focal}
Lin, T.-Y.; Goyal, P.; Girshick, R.; He, K.; and Doll{\'a}r, P. 2017.
\newblock Focal loss for dense object detection.
\newblock In \emph{International Conference on Computer Vision}, 2980--2988.

\bibitem[{Lin et~al.(2014)Lin, Maire, Belongie, Hays, Perona, Ramanan,
  Doll{\'a}r, and Zitnick}]{lin2014microsoft}
Lin, T.-Y.; Maire, M.; Belongie, S.; Hays, J.; Perona, P.; Ramanan, D.;
  Doll{\'a}r, P.; and Zitnick, C.~L. 2014.
\newblock Microsoft coco: Common objects in context.
\newblock In \emph{European Conference on Computer Vision}, 740--755.

\bibitem[{Liu et~al.(2019)Liu, Yang, Liu, Song, Li, and Chen}]{Liu2019Dpatch}
Liu, X.; Yang, H.; Liu, Z.; Song, L.; Li, H.; and Chen, Y. 2019.
\newblock {Dpatch: An adversarial patch attack on object detectors}.
\newblock In \emph{AAAI Workshop on Artificial Intelligence Safety}.

\bibitem[{Liu et~al.(2017)Liu, Chen, Liu, and Song}]{Liu2017Delving}
Liu, Y.; Chen, X.; Liu, C.; and Song, D. 2017.
\newblock {Delving into transferable adversarial examples and black-box
  attacks}.
\newblock \emph{International Conference on Learning Representations}.

\bibitem[{Liu et~al.(2018)Liu, Wang, Shan, and Chen}]{liu2018structure}
Liu, Y.; Wang, R.; Shan, S.; and Chen, X. 2018.
\newblock Structure inference net: Object detection using scene-level context
  and instance-level relationships.
\newblock In \emph{IEEE Conference on Computer Vision and Pattern Recognition},
  6985--6994.

\bibitem[{Madry et~al.(2018)Madry, Makelov, Schmidt, Tsipras, and
  Vladu}]{madry2017towards}
Madry, A.; Makelov, A.; Schmidt, L.; Tsipras, D.; and Vladu, A. 2018.
\newblock Towards deep learning models resistant to adversarial attacks.
\newblock In \emph{International Conference on Learning Representations}.

\bibitem[{Marques, Barenholtz, and Charvillat(2011)}]{marques2011context}
Marques, O.; Barenholtz, E.; and Charvillat, V. 2011.
\newblock Context modeling in computer vision: techniques, implications, and
  applications.
\newblock \emph{Multimedia Tools and Applications}, 51(1): 303--339.

\bibitem[{Moosavi-Dezfooli, Fawzi, and Frossard(2016)}]{moosavi2016deepfool}
Moosavi-Dezfooli, S.-M.; Fawzi, A.; and Frossard, P. 2016.
\newblock {Deepfool: A Simple and Accurate Method to Fool Deep Neural
  Networks}.
\newblock In \emph{IEEE Conference on Computer Vision and Pattern Recognition},
  2574--2582.

\bibitem[{Mottaghi et~al.(2014)Mottaghi, Chen, Liu, Cho, Lee, Fidler, Urtasun,
  and Yuille}]{Mottaghi2014ConWild}
Mottaghi, R.; Chen, X.; Liu, X.; Cho, N.~G.; Lee, S.~W.; Fidler, S.; Urtasun,
  R.; and Yuille, A. 2014.
\newblock {The role of context for object detection and semantic segmentation
  in the wild}.
\newblock In \emph{IEEE Conference on Computer Vision and Pattern Recognition},
  891--898.

\bibitem[{Pang et~al.(2021)Pang, Chen, Li, Xu, Feng, Shi, Ouyang, and
  Lin}]{pang2021towards}
Pang, J.; Chen, K.; Li, Q.; Xu, Z.; Feng, H.; Shi, J.; Ouyang, W.; and Lin, D.
  2021.
\newblock Towards Balanced Learning for Instance Recognition.
\newblock \emph{International Journal of Computer Vision}, 129(5): 1376--1393.

\bibitem[{Pang et~al.(2019)Pang, Chen, Shi, Feng, Ouyang, and
  Lin}]{pang2019libra}
Pang, J.; Chen, K.; Shi, J.; Feng, H.; Ouyang, W.; and Lin, D. 2019.
\newblock Libra R-CNN: Towards Balanced Learning for Object Detection.
\newblock In \emph{IEEE Conference on Computer Vision and Pattern Recognition},
  821--830.

\bibitem[{Papernot et~al.(2017)Papernot, McDaniel, Goodfellow, Jha, Celik, and
  Swami}]{Papernot2017Practical}
Papernot, N.; McDaniel, P.; Goodfellow, I.; Jha, S.; Celik, Z.~B.; and Swami,
  A. 2017.
\newblock {Practical black-box attacks against machine learning}.
\newblock In \emph{ACM Asia Conference on Computer and Communications
  Security}.

\bibitem[{Redmon and Farhadi(2018)}]{redmon2018yolov3}
Redmon, J.; and Farhadi, A. 2018.
\newblock Yolov3: An incremental improvement.
\newblock \emph{arXiv preprint arXiv:1804.02767}.

\bibitem[{Ren et~al.(2015)Ren, He, Girshick, and Sun}]{ren2015faster}
Ren, S.; He, K.; Girshick, R.; and Sun, J. 2015.
\newblock Faster r-cnn: Towards real-time object detection with region proposal
  networks.
\newblock In \emph{Advances in Neural Information Processing Systems}, 91--99.

\bibitem[{Saha et~al.(2020)Saha, Subramanya, Patil, and
  Pirsiavash}]{saha2020role}
Saha, A.; Subramanya, A.; Patil, K.; and Pirsiavash, H. 2020.
\newblock Role of spatial context in adversarial robustness for object
  detection.
\newblock In \emph{IEEE Conference on Computer Vision and Pattern Recognition
  Workshops}, 784--785.

\bibitem[{Strat and Fischler(1991)}]{strat1991context}
Strat, T.~M.; and Fischler, M.~A. 1991.
\newblock Context-based vision: recognizing objects using information from both
  2 d and 3 d imagery.
\newblock \emph{IEEE Transactions on Pattern Analysis and Machine
  Intelligence}, 13(10): 1050--1065.

\bibitem[{Torralba, Murphy, and Freeman(2005)}]{torralba2004contextual}
Torralba, A.; Murphy, K.~P.; and Freeman, W. 2005.
\newblock Contextual Models for Object Detection Using Boosted Random Fields.
\newblock In \emph{Advances in Neural Information Processing Systems},
  volume~17, 1401–--1408.

\bibitem[{Wang et~al.(2020{\natexlab{a}})Wang, Sun, Kortylewski, and
  Yuille}]{wang2020robust}
Wang, A.; Sun, Y.; Kortylewski, A.; and Yuille, A.~L. 2020{\natexlab{a}}.
\newblock Robust object detection under occlusion with context-aware
  compositionalnets.
\newblock In \emph{IEEE Conference on Computer Vision and Pattern Recognition},
  12645--12654.

\bibitem[{Wang and He(2021)}]{wang2021enhancing}
Wang, X.; and He, K. 2021.
\newblock Enhancing the Transferability of Adversarial Attacks through Variance
  Tuning.
\newblock In \emph{IEEE Conference on Computer Vision and Pattern Recognition},
  1924--1933.

\bibitem[{Wang et~al.(2020{\natexlab{b}})Wang, Tan, Zhang, Zhao, and
  Kuang}]{Wang2020blackobject}
Wang, Y.; Tan, Y.; Zhang, W.; Zhao, Y.; and Kuang, X. 2020{\natexlab{b}}.
\newblock {An adversarial attack on DNN-based black-box object detectors}.
\newblock \emph{Journal of Network and Computer Applications}, 161: 102634.

\bibitem[{Wei et~al.(2019)Wei, Liang, Chen, and Cao}]{Wei2019Transferable}
Wei, X.; Liang, S.; Chen, N.; and Cao, X. 2019.
\newblock Transferable Adversarial Attacks for Image and Video Object
  Detection.
\newblock In \emph{International Joint Conference on Artificial Intelligence},
  954--960.

\bibitem[{Wu et~al.(2020)Wu, Lim, Davis, and Goldstein}]{wu2020making}
Wu, Z.; Lim, S.-N.; Davis, L.~S.; and Goldstein, T. 2020.
\newblock Making an invisibility cloak: Real world adversarial attacks on
  object detectors.
\newblock In \emph{European Conference on Computer Vision}, 1--17.

\bibitem[{Xie et~al.(2017)Xie, Wang, Zhang, Zhou, Xie, and
  Yuille}]{xie2017adversarial}
Xie, C.; Wang, J.; Zhang, Z.; Zhou, Y.; Xie, L.; and Yuille, A. 2017.
\newblock Adversarial examples for semantic segmentation and object detection.
\newblock In \emph{International Conference on Computer Vision}, 1369--1378.

\bibitem[{Xie et~al.(2019)Xie, Zhang, Zhou, Bai, Wang, Ren, and
  Yuille}]{xie2019improving}
Xie, C.; Zhang, Z.; Zhou, Y.; Bai, S.; Wang, J.; Ren, Z.; and Yuille, A.~L.
  2019.
\newblock Improving transferability of adversarial examples with input
  diversity.
\newblock In \emph{IEEE Conference on Computer Vision and Pattern Recognition},
  2730--2739.

\bibitem[{Yao, Fidler, and Urtasun(2012)}]{Yao2012WholeScene}
Yao, J.; Fidler, S.; and Urtasun, R. 2012.
\newblock {Describing the scene as a whole: Joint object detection, scene
  classification and semantic segmentation}.
\newblock \emph{IEEE Conference on Computer Vision and Pattern Recognition},
  702--709.

\bibitem[{Yin et~al.(2021)Yin, Li, Cai, Song, Asif, Roy-Chowdhury, and
  Krishnamurthy}]{yin2021exploiting}
Yin, M.; Li, S.; Cai, Z.; Song, C.; Asif, M.~S.; Roy-Chowdhury, A.~K.; and
  Krishnamurthy, S.~V. 2021.
\newblock Exploiting multi-object relationships for detecting adversarial
  attacks in complex scenes.
\newblock In \emph{International Conference on Computer Vision}, 7858--7867.

\bibitem[{Zhang, Zhou, and Li(2020)}]{Zhang2020CAA}
Zhang, H.; Zhou, W.; and Li, H. 2020.
\newblock {Contextual Adversarial Attacks for Object Detection}.
\newblock \emph{International Conference on Multimedia and Expo}, 1--6.

\bibitem[{Zhang et~al.(2017)Zhang, Elhoseiny, Cohen, Chang, and
  Elgammal}]{zhang2017relationship}
Zhang, J.; Elhoseiny, M.; Cohen, S.; Chang, W.; and Elgammal, A. 2017.
\newblock Relationship proposal networks.
\newblock In \emph{IEEE Conference on Computer Vision and Pattern Recognition},
  5678--5686.

\bibitem[{Zhang et~al.(2019)Zhang, Wan, Liu, Ji, and Ye}]{zhang2019freeanchor}
Zhang, X.; Wan, F.; Liu, C.; Ji, R.; and Ye, Q. 2019.
\newblock {FreeAnchor}: Learning to Match Anchors for Visual Object Detection.
\newblock In \emph{Neural Information Processing Systems}, 147--155.

\bibitem[{Zhu et~al.(2021)Zhu, Su, Lu, Li, Wang, and Dai}]{zhu2021deformable}
Zhu, X.; Su, W.; Lu, L.; Li, B.; Wang, X.; and Dai, J. 2021.
\newblock Deformable DETR: Deformable Transformers for End-to-End Object
  Detection.
\newblock In \emph{International Conference on Learning Representations}.

\end{thebibliography}
\newpage
\clearpage
\label{sec:appendix}
\onecolumn
\noindent \begin{center} {\LARGE  \textbf{Supplementary Material}} \end{center}
\setcounter{section}{0}
\renewcommand\thesection{\Alph{section}}


\section{More Analysis on Number of Helper Objects}
We present additional results for perturbation levels $L_\infty \le 10, 30.$ We observe a similar trend as in Figure~\ref{fig:analysis_study_mis} that success of mis-categorization attacks increases as we add helper objects in our attack plans. In some cases, the success rate almost doubles compared to baseline as we add 5 helper objects. 

\begin{figure*}[!ht]
    \centering 
    \begin{subfigure}[t]{0.45\linewidth}
    \includegraphics[width=1\linewidth]{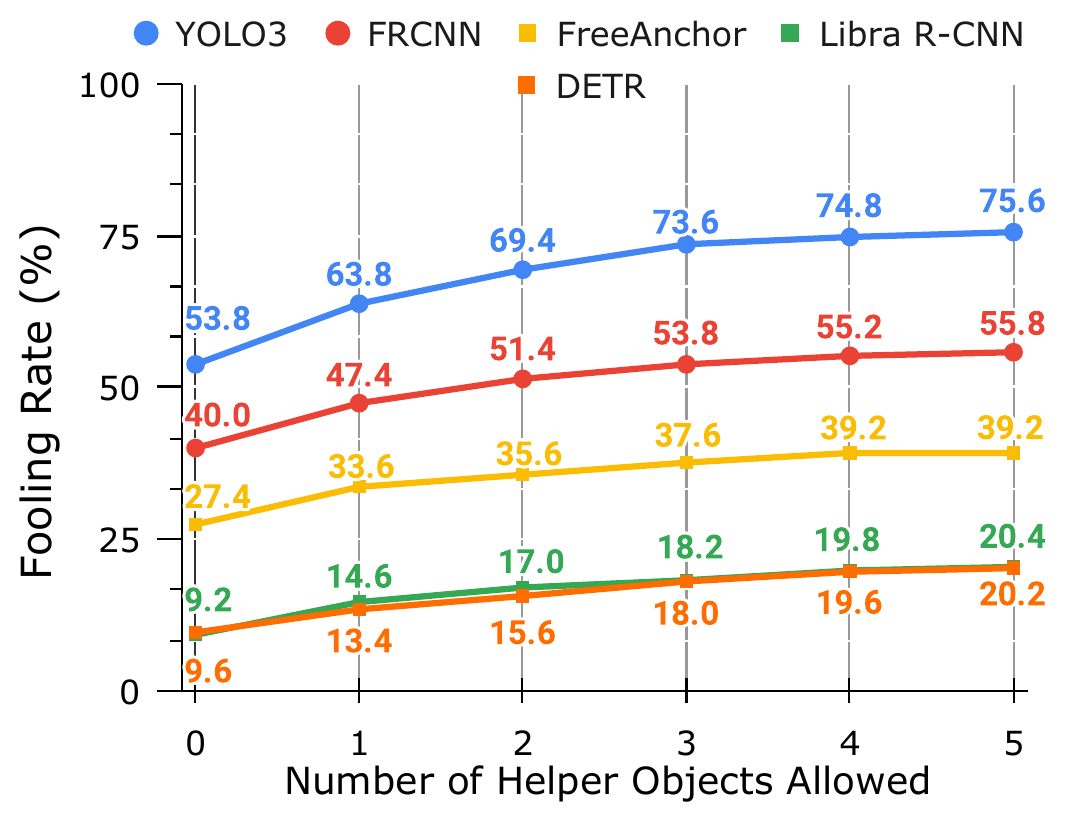}
    \caption{PASCAL VOC, $L_{\infty} \le 10$}
    \end{subfigure}
    \begin{subfigure}[t]{0.45\linewidth}
    \includegraphics[width=1\textwidth]{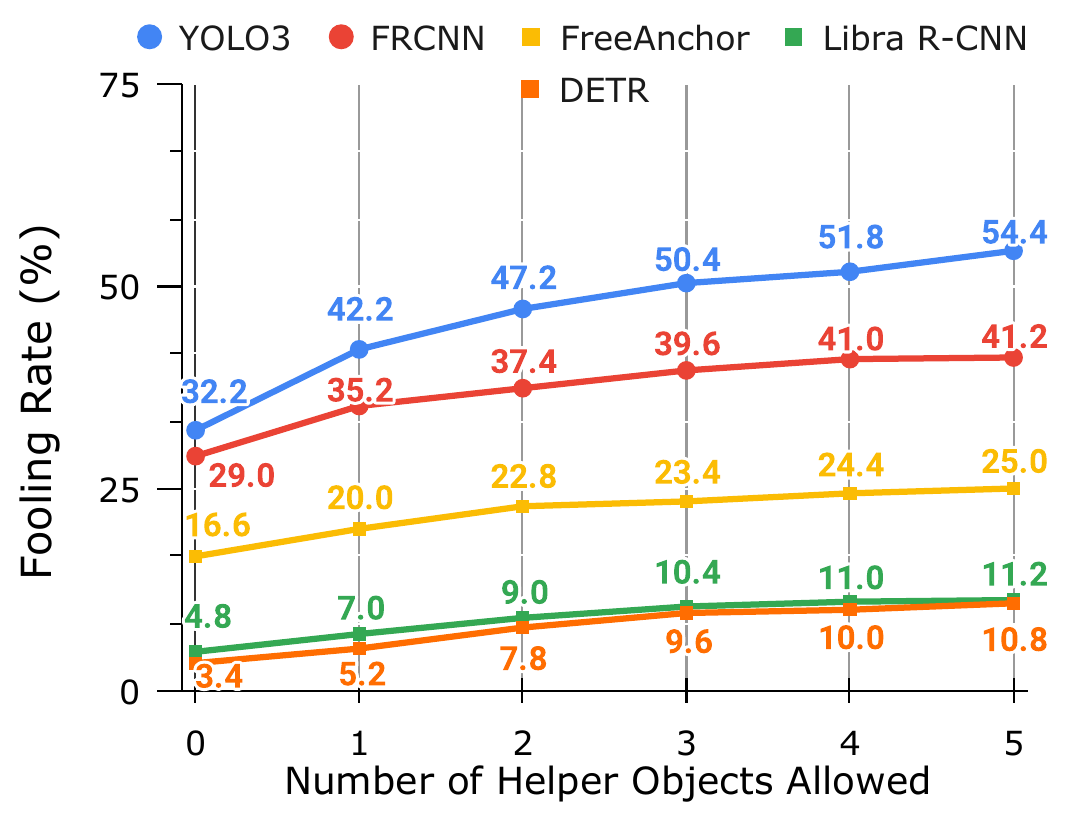}
    \caption{MS COCO, $L_{\infty} \le 10$}
    \end{subfigure}
    \par\bigskip
    \begin{subfigure}[t]{0.45\linewidth}
    \includegraphics[width=1\linewidth]{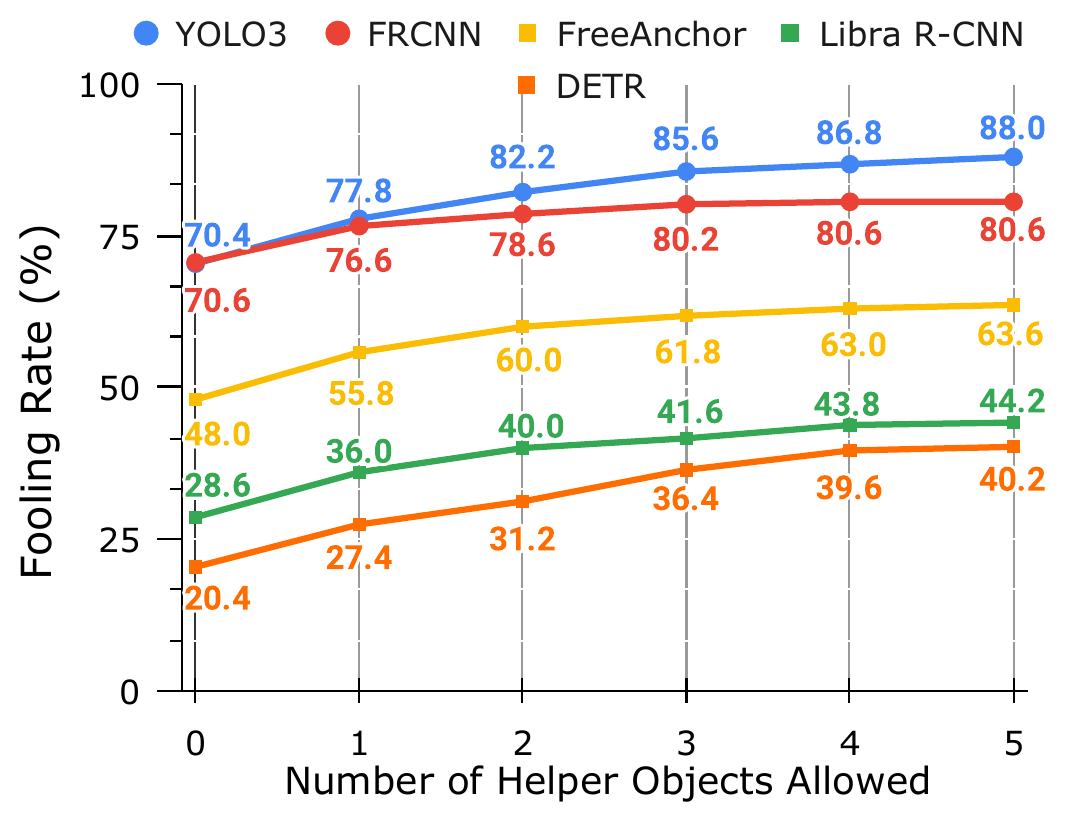}
    \caption{PASCAL VOC, $L_{\infty} \le 30$}
    \end{subfigure}
    \begin{subfigure}[t]{0.45\linewidth}
    \includegraphics[width=1\textwidth]{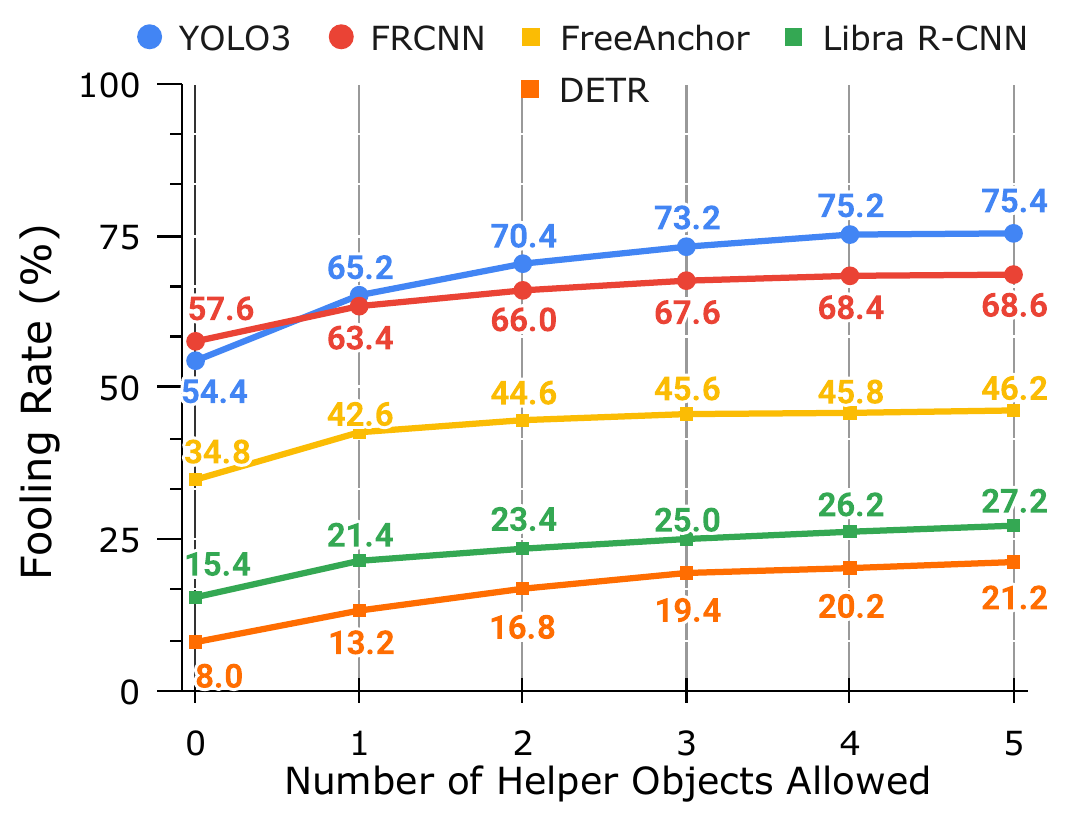}
    \caption{MS COCO, $L_{\infty} \le 30$}
    \end{subfigure}
    \caption{Mis-categorization attack fooling rate of white-box and black-box models at perturbation level $L_{\infty} \le 10,30$ w.r.t. number of helper objects allowed (changed or added). In the legend, circle denotes white-box models (FRCNN and YOLO3) and square denotes black-box models (FreeAnchor, Libra R-CNN, and DETR). Baseline is where no helper objects is allowed.}
    \label{fig:analysis_study_mis_supp}
\end{figure*}

\section{More Visualization Examples}
We present some additional images to show comparison between our context-aware attack method with baseline method. We show examples where the perturbations generated by our method can successfully transfer to the blackbox model while the perturbations generated by baseline method fail. The experiment settings are the same as Figure 2 in the main paper.

\begin{figure*}[!ht]
    \centering 
    \begin{subfigure}[t]{0.9\linewidth}
    \includegraphics[width=1\linewidth]{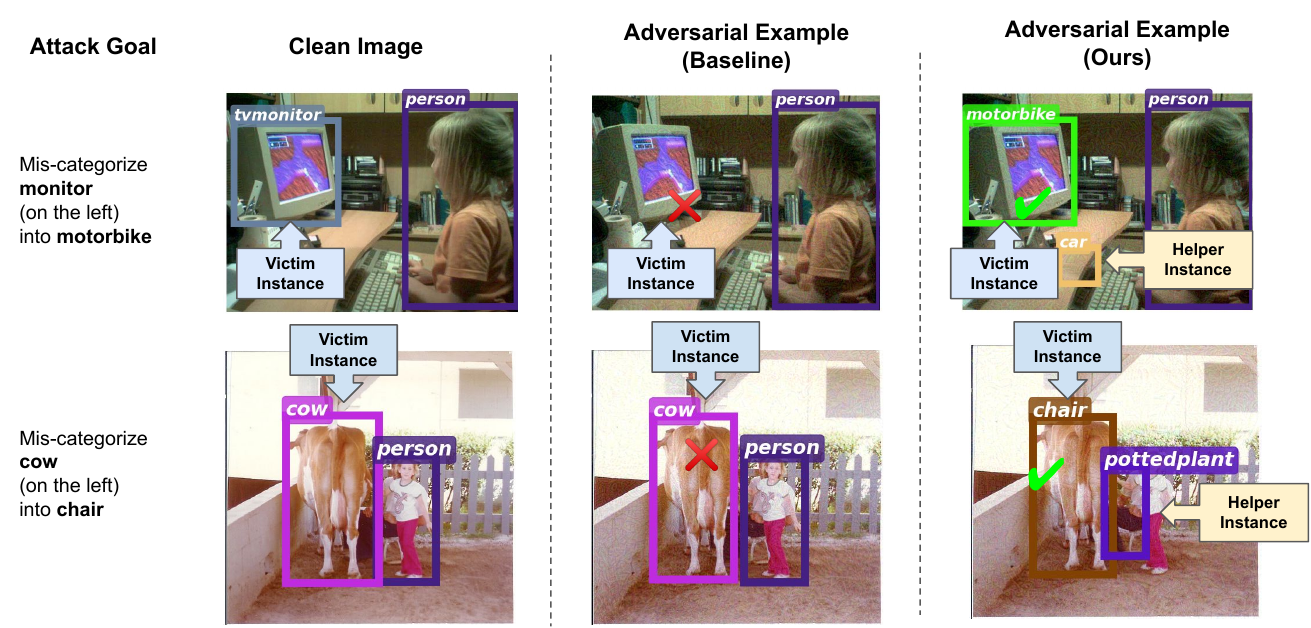}
    \caption{PASCAL VOC, $L_{\infty} \le 20$}
    \end{subfigure}
    \par\bigskip
    \begin{subfigure}[t]{0.9\linewidth}
    \includegraphics[width=1\textwidth]{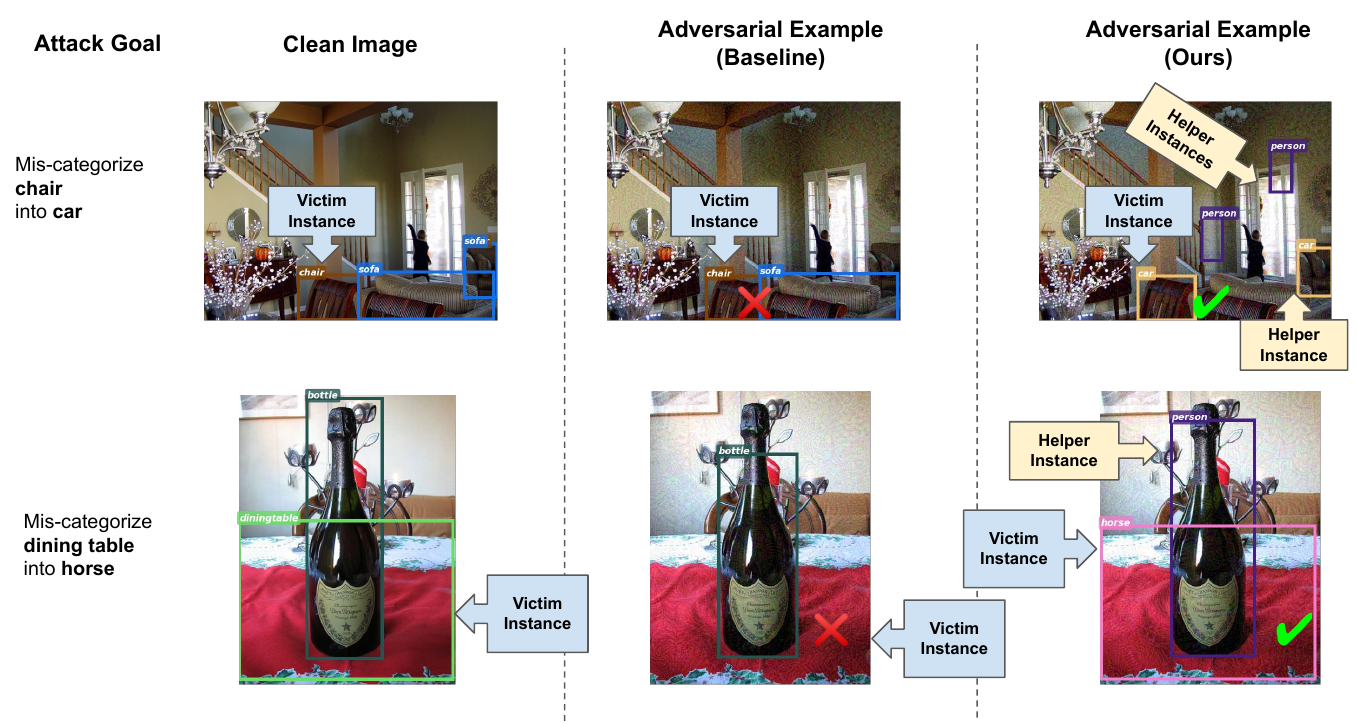}
    \caption{PASCAL VOC, $L_{\infty} \le 30$}
    \end{subfigure}
    \caption{Supplement to Figure 2, here we visualize four more examples under different perturbation budgets ($L_{\infty} \le 20,30$) where baseline attack fails but our context-aware method succeeds by introducing helper objects in the attack. The perturbation is generated from our perturbation machine (whitebox ensemble of FRCNN and YOLOv3) and tested on the blackbox model (RetinaNet). The detection results on original image, image perturbed by baseline attack, and image perturbed by our context-aware method are shown in the subfigures from left to right. In these examples, we introduce car as a helper object to mis-categorize the victim monitor to motorbike, introduce a potted plant to mis-categorize the cow to a chair, add a few persons and a car to mis-categorize the chair to a car, and change the bottle to a person in order to mis-categorize the dining table to a horse.}
    \label{fig:visualization_supp_voc}
\end{figure*}
\begin{figure*}[!ht]
    \centering 
    \begin{subfigure}[t]{0.75\linewidth}
    \includegraphics[width=1\linewidth]{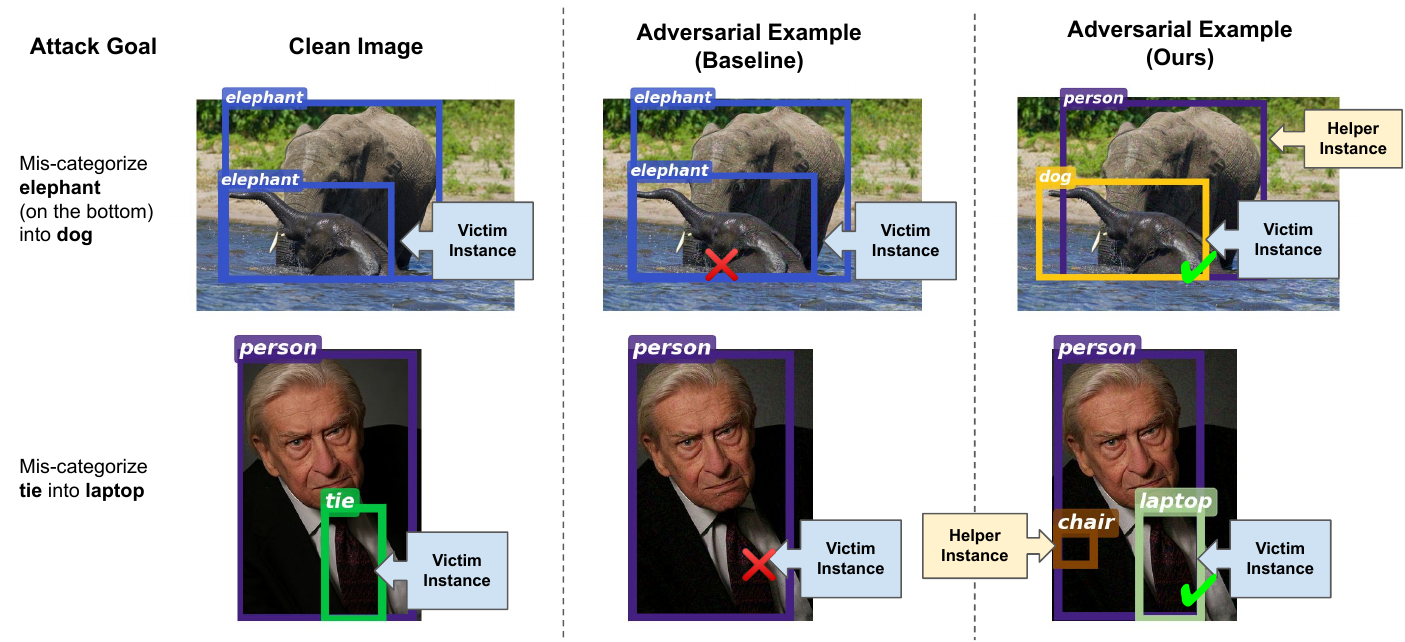}
    \caption{MS COCO, $L_{\infty} \le 10$}
    \end{subfigure}
    \begin{subfigure}[t]{0.75\linewidth}
    \includegraphics[width=1\textwidth]{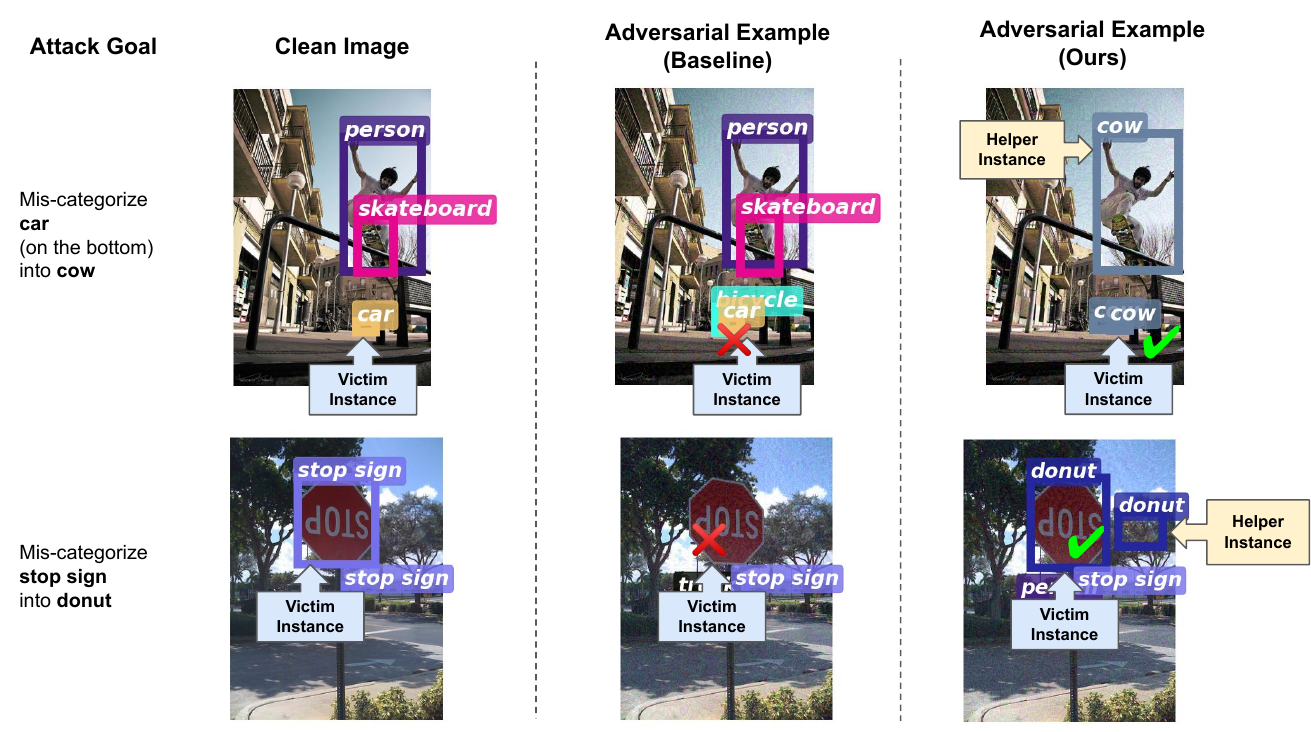}
    \caption{MS COCO, $L_{\infty} \le 20$}
    \end{subfigure}
    \begin{subfigure}[t]{0.75\linewidth}
    \includegraphics[width=1\textwidth]{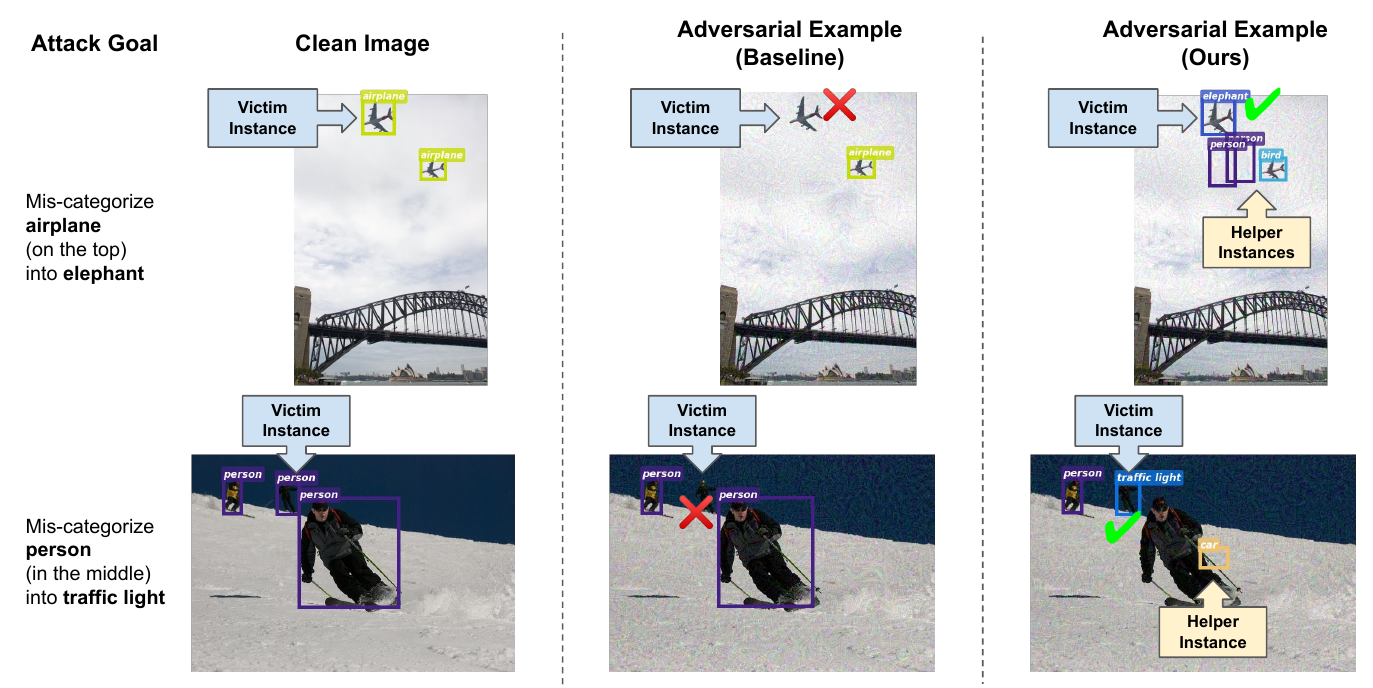}
    \caption{MS COCO, $L_{\infty} \le 30$}
    \end{subfigure}
    \caption{\footnotesize Correspond to the previous visualizations on VOC dataset, here we also visualize examples for COCO dataset, where baseline attack fails but our context-aware method succeeds by introducing helper objects in the attack. The perturbation ($L_{\infty} \le 10,20,30$) is generated from our perturbation machine (whitebox ensemble of FRCNN and YOLOv3) and tested on the blackbox model (RetinaNet). The detection results on original image, image perturbed by baseline attack, and image perturbed by our context-aware method are shown in the subfigures from left to right. In (a), we introduce a person as a helper object to mis-categorize the victim elephant to a dog, introduce a chair to mis-categorize the tie to a laptop; in (b), we add a few cows in the scene to mis-categorize the car to a cow, added an other donut to mis-categorize the stop sign to a donut; in (c), we perturb the airplane a bird and add a few persons to mis-categorize the airplane to an elephant, introduce a car to mis-categorize the person to a traffic light.}
    \label{fig:visualization_supp_coco}
\end{figure*}

\end{document}